\documentclass[preprint]{sig-alternate-05-2015}
\usepackage{color}
\usepackage{multirow}
\usepackage{placeins}
\usepackage{afterpage}
\usepackage{subcaption}

\begin{document}


%
%
%
\newcommand{\red}[1]{\protect\begin{color}{red}#1\protect\end{color}}
\tolerance 4000  
%
%
%
%

%
\toappearbox{This is the pre-print version of the paper set to appear as a part of Multimedia Grand Challenge Solutions track in ACMMM '16, October 15-19, 2016, Amsterdam, Netherlands}



\title{%
Frame- and Segment-Level Features and Candidate Pool \\[1mm] Evaluation for Video Caption Generation
}

%
%
%
%

 \numberofauthors{2}
 \author{
  \alignauthor
  Rakshith Shetty\\
  \affaddr{Dept.\@ of Computer Science}\\
  \affaddr{Aalto University, Espoo, Finland}\\
  \email{rakshith.shetty@aalto.fi}
  \alignauthor
  Jorma Laaksonen\\
  \affaddr{Dept.\@ of Computer Science}\\
  \affaddr{Aalto University, Espoo, Finland}\\
  \email{jorma.laaksonen@aalto.fi}
}

\maketitle


\begin{abstract}

  We present our submission to the Microsoft Video to Language
  Challenge of generating short captions describing videos in the
  challenge dataset.
  Our model is based on the encoder--decoder pipeline, popular in
  image and video captioning systems.
  We propose to utilize two different kinds of video features, one to
  capture the video content in terms of objects and attributes, and
  the other to capture the motion and action information.
  Using these diverse features we train models specializing in two
  separate input sub-domains.
  We then train an evaluator model which is used to pick the best caption
  from the pool of candidates generated by these domain expert models.
  We argue that this approach is better suited for the current video
  captioning task, compared to using a single model, due to the
  diversity in the dataset.

  Efficacy of our method is proven by the fact that it was rated best
  in MSR Video to Language Challenge, as per human
  evaluation.
  Additionally, we were ranked second in the automatic evaluation
  metrics based table.

\end{abstract}

\section{Introduction}

The problem of describing videos using natural language has garnered a
lot of interest after the great progress recently made in image
captioning.
This development has been partly driven also by the availability of
large datasets of images and videos with human-annotated
captions describing them.
The M-VAD~\cite{rohrbach15cvpr} and MPII-MD~\cite{AtorabiM-VAD2015}
datasets used in the first Large Scale Movie Description
Challenge suffered from having only one reference 
caption for each video.
Multiple reference captions improve the language models learned and
also lead to better automatic evaluation as shown
in~\cite{Vedantam_2015_CVPR}.
The Microsoft Video to Language dataset~(MSR-VTT)~\cite{Xu:CVPR16}
addresses this by providing 20 captions for each video.

A popular recipe for solving the visual captioning problem has been to
use an encoder--decoder model~\cite{Vinyals_2015_CVPR,
  venugopalan2015sequence}.
The encoder produces a feature vector representation of the
input, which the decoder, usually a Long-Short Term
Memory~(LSTM) network, takes as input and
generates a caption.

Unlike in the case of images, where the convolutional neural network
(CNN) image features have become the \emph{de facto} standard features
for many image understanding related tasks, no single video feature
extraction method has achieved the best performance across tasks and
datasets.
Dense trajectories~\cite{DBLP:conf/cvpr/WangKSL11} and Improved dense
trajectories~\cite{Wang2013} have been popular hand-crafted video
feature extractors in action recognition.
Following the success of deep CNNs on static images, there have
been attempts with 3-D CNNs which operate directly on video
segments~\cite{KarpathyCVPR14, DBLP:C3D}.
However, these models need a lot of training data and are usually
pre-trained on some large dataset, e.g.\@ the Sports-1M data.

The task of caption generation also requires us to describe
the objects seen in the video and their attributes, in addition to
recognizing actions.
This can be solved by extracting features from individual
frames~\cite{venugopalan2015sequence} or
keyframes~\cite{shetty2015video} using CNNs pre-trained on the
ImageNet dataset.

In this work we explore both problems, video feature extraction and
visual content description, for the task of automatic video
captioning.
We experiment with methods which treat videos as bags of images and
with methods mainly focusing on capturing the motion information
present in videos.
We also study how to combine these features and present an effective
method to ensemble multiple caption generators trained on different
video features.

%
%
%
%


\section{Dataset}

The MSR-VTT dataset consists of 10000 video clips with 20
human-annotated captions for each of them.
Each video belongs to one of 20 categories including \emph{music},
\emph{gaming}, \emph{sports}, \emph{news}, etc.
The dataset is split into training~(6513 clips), validation~(497
clips) and test~(2990 clips) sets. 
The training set videos are between 10 to 30 seconds long, with most
of them less than 20 seconds in duration.

The reference captions come from a vocabulary of $\sim$30k words,
which we have filtered down to $\sim$8k after removing words occurring
fewer than 5 times.
Although the number of video clips in the dataset is relatively small
compared to M-VAD~\cite{rohrbach15cvpr} and
MPII-MD~\cite{AtorabiM-VAD2015}, the dataset is attractive due to the
diversity of its videos and due to the larger number of reference
captions per video.

Any video captioning algorithm applied to this dataset needs to handle
very diverse video styles, eg.\@ video game play footage, news and
interviews.
Action recognition based features will suffer with videos with lots of
cuts and scene changes.
Conversely, with frame-based features, the system will fail to
identify fine-grained differences in diverse action-oriented
categories like sports or cooking.
Thus our approach is to use multiple captioning models, each trained
on different types of features, and an evaluator network to pick the
final caption among the candidates.


\section{Our Method}


\subsection{Overview}

We use the standard encoder--decoder framework for our video caption
generator.
This framework has been widely used in the literature for machine
translation and image~\cite{Vinyals_2015_CVPR} and
video~\cite{shetty2015video, DBLP:journals/corr/RohrbachTRTPLCS16}
captioning.
The encoder stage consists of various feature extraction modules which
encode the input video into a vectorial form.
The decoder is the language model that generates a caption using these
feature vectors.

We also utilize an additional model, an evaluator network which takes
as its inputs a generated caption and a video feature, and computes a
score measuring how well these two match.
This evaluator is used to build an ensemble of multiple caption
generator models trained on different features.


\subsection{Feature Extraction}
\label{subsec-featExtract}

We use two different paradigms for video feature extraction.
The first one is to treat the video as just a sequence of 2-D static
images and use CNNs trained on ImageNet~\cite{imagenet_cvpr09} to
extract static image features from these frames.
The second approach is to treat the video as 3-D data, consisting of a
sequence of video segments, and use methods which also consider the
variations along the time dimension for feature extraction.
In both the above methods, pooling methods are used to combine
multiple frame/segment-level features into one video-level feature
vector.


\subsubsection{Frame-Level Feature Extraction}

We use the GoogLeNet~\cite{DBLP:journals/corr/SzegedyLJSRAEVR14} model
trained on ImageNet for extracting frame-level features.
For efficiency, we only extract these features from one frame every
second. 

In the GoogLeNet we have used the \emph{5th Inception module}, having
the dimensionality of 1024.
We augment these features with the reverse spatial pyramid pooling
proposed in~\cite{Gong2014} with two scale levels.
The second level consists of a $3\times3$ grid with overlaps and
horizontal flipping, resulting in a total of 26 regions, on the scale
of two.
The activations of the regions are then pooled using diverse
combinations of average and maximum pooling.
Finally, the activations of the different scales are concatenated
resulting in 2048-dimensional features.
By using the varying combinations of pooling techniques, we have
obtained three somewhat different feature sets from the same network.

We use the mean pooling to fuse the multiple frame-level
features extracted from one video into a single vector.


\subsubsection{Video Segment Feature Extraction}

For extracting segment-based features we have used two different
algorithms, namely dense trajectories~\cite{DBLP:conf/cvpr/WangKSL11,
Wang2013} and 3-D CNN network based C3D features~\cite{DBLP:C3D}.

We have used both the standard~\cite{DBLP:conf/cvpr/WangKSL11} and
the improved versions~\cite{Wang2013} of the dense trajectory
features.
For both versions, the trajectories and their descriptors (HOG, HOF,
MBHx and MBHy as described in~\cite{DBLP:conf/cvpr/WangKSL11}) are
first extracted for the entire video.
All of these five features are separately encoded into a fixed-size
vector using bag-of-features encoding with a codebook of 1000 vectors.
Each codebook is obtained using k-means clustering on 250k random 
trajectory samples from the training set.
Finally, concatenating the vector encodings of each of the descriptors
we get a video feature vector of 5000 dimensions. 

As an alternative, we also extract video-segment features with the
deep 3-D convolutional model, C3D~\cite{DBLP:C3D}, pre-trained on the
Sports-1M dataset.
The features are extracted for every 16 frames from the \emph{fc7}
layer of the network and mean pooled to get a 4096-dimensional feature
vector representation of the video.

We also utilize the video category information available
for all videos in all splits of the dataset.
This is input to the language model as a one-hot vector of
20 dimensions.


\subsection{Language Model}

\begin{figure}[tb] 
  \centering
  \includegraphics[width=0.3\textwidth]{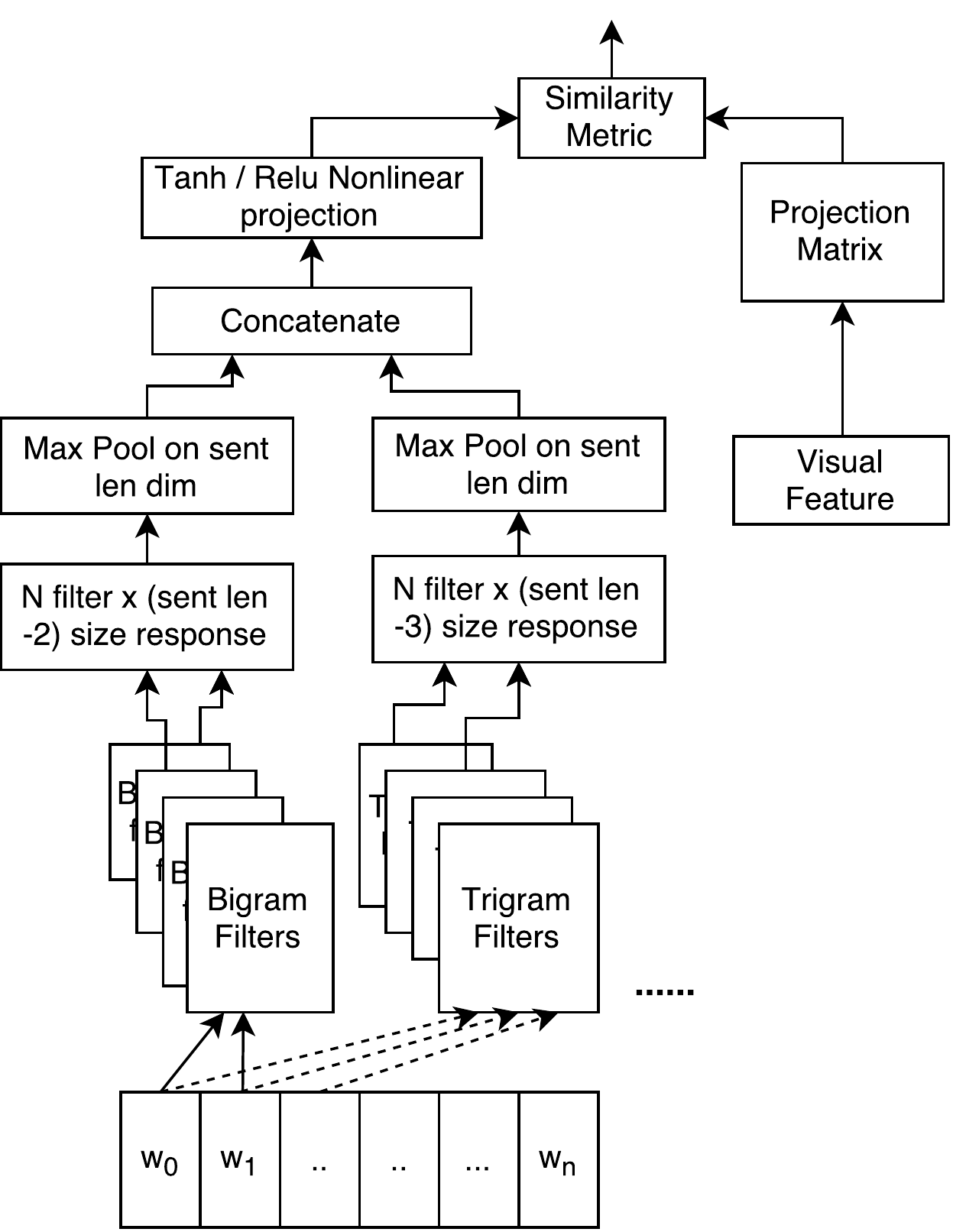} 
  \caption{CNN-based evaluator network to compute the similarity between 
  a video and a caption.}
  \label{fig:CNNEval} 
\end{figure}

\begin{table*}[tbh]
  \caption{Performance of various features and 
    network depths on the validation set of MSR-VTT}
  \vspace{-2mm}
  \newcommand{\modpar}[4]{%
    \multirow{2}{*}{\emph{#1}} & \multirow{2}{*}{#2} & \multirow{2}{*}{#3}
    & \multirow{2}{*}{#4}}
  \centering
  \newcommand{\bs}{\small\bf}
  \begin{tabular}{||c|c|c|c|c|c|c|c|c|}
    \hline\hline
    \bf\# &\bf init &\bf persist &\bf depth &\bf perplex &\bs BLEU-4 &\bs METEOR &\bs CIDEr &\bs ROUGE-L \\\hline\hline
    1 & dt  & gCNN+20Categ & 1  & 27.31 & 0.396 & 0.268 & 0.438 & 0.588 \\
    2 & dt  & gCNN+20Categ & 2  & 27.73 & 0.409 & 0.268 & 0.433 & 0.598 \\
    3 & dt  & gCNN+20Categ & 3  & 28.44 & 0.370 & 0.262 & 0.397 & 0.575 \\\hline
    4 & idt & gCNN+20Categ & 2  & 28.13 & 0.398 & 0.268 & 0.432 & 0.587 \\
    5 & dt  & c3dfc7       & 2  & 29.58 & 0.369 & 0.268 & 0.413 & 0.577 \\\hline
    6 & \multicolumn{4}{c|}{\em CNN evaluator based ensemble of best 4 models}
                                  & \bf0.411 & \bf0.277 & \bf0.464 & \bf0.596 \\\hline
    \hline
  \end{tabular}
  \label{tab:resultsVal}
\end{table*}

We use the LSTM network based language model and training procedures
widely used in visual captioning~\cite{Vinyals_2015_CVPR} with
some tweaks.
Firstly, we train deeper LSTM networks with residual connections
between the layers.
This is similar to the residual connections introduced
in~\cite{He2015} for CNNs.
In our case, the output of the lower LSTM layer is added to the
output of the above layer with residual connections.

We also allow the language model to use two separate input features,
one to initialize the network and the other one to persist throughout
the caption generation process as shown in~\cite{shetty2015video}.
Thus our language model can be trained using two different feature
inputs, \emph{init} and \emph{persist}, fed through separate channels.
We utilize this architecture to train many different language models
by using the diverse feature pairs from the ones described in
Section~\ref{subsec-featExtract}.


\subsection{Evaluator Network}

Using the different video feature combinations we have, we can
train diverse caption generator models. 
Examining the captions generated by them for a set of videos, we find
that each model tends to generate the best captions for different
videos.
If we can reliably evaluate the suitability of a caption for a
given video, we can pick out the best candidate and achieve better
results than with any single model.

We implement this by training a new \emph{evaluator network} whose
task is to pick out the best candidate from the candidate set, given
an input video.
The model takes as input one video feature and one input sentence and
computes a similarity metric between them. 
The evaluator is trained discriminatively, in contrast to the
language models which have been trained generatively.
Thus, it learns to focus on semantically interesting phrases without
having to learn the details needed to generate a syntactically correct
sentence.

Figure~\ref{fig:CNNEval} shows the block diagram of our evaluator.
It consists of a CNN sentence encoder network which takes a sequence
of word vectors as input and learns a fixed-size vector embedding of
the sentence.
The model also has a projection matrix which maps the video
feature to the same space as the sentence embedding.
We then use the cosine distance to score the similarity between the
two embeddings.


The evaluator model is trained to maximize the score assigned to
matching caption--video pairs, while minimizing the score assigned to
random negative caption--video pairs.
We sample up to 50 negative captions per video from the ground truth
captions of other videos.
The trained evaluator is used to rerank captions from multiple
generator models, trained on different feature combinations and
architectures, to pick the best matching candidate.


\section{Experiments and Results}
\FloatBarrier


All LSTM layers in the language models we train have 512 hidden units,
but we vary the number of layers as shown in Table~\ref{tab:resultsVal}. 
We train the language models on the entire training set of the MSR-VTT
dataset using stochastic gradient descent with the RMSProp algorithm
and dropout regularization.
The errors are backpropagated to all the language model parameters and
word embedding matrices, but the feature extraction models are
kept constant. 

In all our experiments we use beam search to generate sentences from a
trained model.
After experimenting with different beam sizes, we found that the beam
size of $b=5$ works adequately across all our models.



\subsection{Validation Set Results}


\begin{figure*}[thp]
  \begin{center}
  \newcommand{\mcCell}[1]{%
  \multicolumn{1}{c}{#1}}
  \centering
  \begin{tabular}{lll}
    \mcCell{\includegraphics[width=0.25\linewidth]{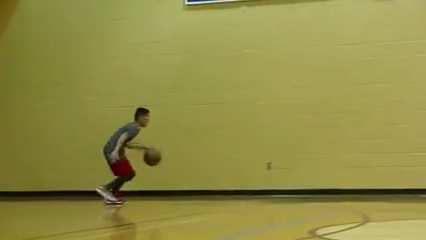}} &
    \mcCell{\includegraphics[width=0.25\linewidth]{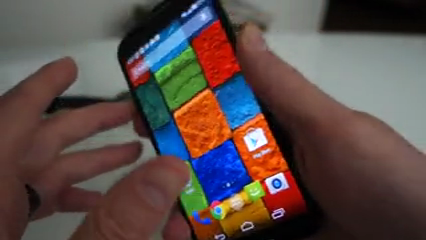}} &
    \mcCell{\includegraphics[width=0.25\linewidth]{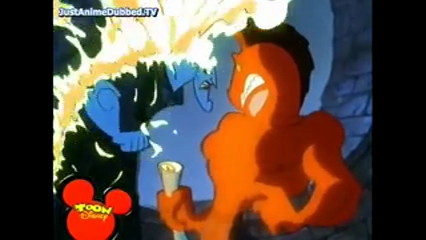}} \\
    \textbf{\em\#6:} a man is running in a gym &
    \textbf{\em\#6:} a person is playing with a rubix cube &
    \textbf{\em\#6:} cartoon characters are interacting\\
    \textbf{\em\#2:} a man is running&
    \textbf{\em\#2:} a man is holding a phone&
    \textbf{\em\#2:} a person is playing a video game\\
    \textbf{\em\#5:} a man is playing basketball&
    \textbf{\em\#5:} a person is playing with a rubix cube &
    \textbf{\em\#5:} a group of people are talking\\
  \end{tabular}
  \end{center}
  \vspace{-5mm}
  \caption{Sample captions generated for some test set videos by our
          models.}
  \label{fig:capSamps}
\end{figure*}

\begin{table*}[hbt]
  \caption{Top 5 teams as per metrics and human evaluations on the test set}
  \vspace{-2mm}
  \newcommand{\bs}{\small\bf}
  \centering
  \begin{subtable}{.5\linewidth}
  \centering
  \begin{tabular}{||c|c|c|c|c|}
    \hline\hline
    \bf Team  &\bs BLEU-4 &\bs METEOR &\bs CIDEr &\bs ROUGE-L \\\hline\hline
    v2t\_navigator &\bf0.408 &\bf0.282 & 0.448 &\bf0.609 \\
    \bf Aalto      & 0.398 & 0.269 &0.457 & 0.598 \\
    VideoLAB       & 0.391 & 0.277 & 0.441 & 0.606 \\
    ruc-uva        & 0.387 & 0.269 &\bf0.459 & 0.587 \\
    Fudan-ILC      & 0.387 & 0.268 & 0.419 & 0.595 \\\hline
    \hline
  \end{tabular}
  \caption{Automatic metric-based evaluation}%
  \label{tab:resultsTestMet}
  \end{subtable}%
  \begin{subtable}{.5\linewidth}
  \centering
  \begin{tabular}{||c|c|c|c|}
    \hline\hline
    \bf Team  &\bs C1 &\bs C2 &\bs C3 \\\hline\hline
    \bf Aalto      & \bf3.263 & 3.104 & \bf3.244\\
    v2t\_navigator & 3.261 & 3.091 & 3.154 \\
    VideoLAB       & 3.237 & \bf3.109 & 3.143 \\
    Fudan-ILC      & 3.185 & 2.999 & 2.979 \\
    ruc-uva        & 3.225 & 2.997 & 2.933 \\\hline
    \hline
  \end{tabular}
  \caption{Human evaluation}
  \label{tab:resultsTestHum}
  \end{subtable}
  \vspace{-5mm}
\end{table*}

In order to measure the performance differences due to the different
feature combinations and architectural changes, we use the validation
set of the MSR-VTT dataset which contains 497 videos.
Performance is measured quantitatively using the standard automatic
evaluation metrics, namely METEOR, CIDEr, ROUGE-L and BLEU. 

Table~\ref{tab:resultsVal} shows the results on the validation set.
The columns \emph{init} and \emph{persist} indicate the features used
for those input channels, respectively, in the language model.
The column \emph{depth} is the number of layers in the language
model and \emph{perplex} is the perplexity measure on the validation
set.

Models \#1, \#2 and \#3 all use the dense trajectory (dt) features as
\emph{init} input and the mean pooled frame-level GoogLeNet features
concatenated with the video category vector (gCNN+20Categ) as the
\emph{persist} input.
They vary in the number of layers in the language model.
Comparing their performances we see that the 2-layer model outperforms
the single layered model by a small margin, while the 3-layer version
is the inferior one.

Model \#4 is similar to \#2, but uses the improved dense trajectories
(idt) as the \emph{init} input instead.
Model \#5 differs from \#2 by the fact that it uses mean pooled 3-D
convolutional features as the \emph{persist} input.
We see that both \#4 and \#5 are competitive, but slightly worse
than our best single model, \#2.
Upon qualitatively analyzing the model outputs, we see that each of
them performs well on different kinds of videos.
For example, model \#5, which only uses input features trained for action
recognition, does well in videos involving a lot of motion, but suffers
in recognizing the overall scenery of the video.
Conversely, model \#2 trained on frame-level features does better in
recognizing objects and scenes, but makes mistakes with the sequence
of their appearance, possibly due to the pooling operation.
This fact can be observed in examples in Figure~\ref{fig:capSamps}.
Model \#5 produces a better caption on the video in the first column,
while \#2 does better on the video in the second column.

To get maximum utility out of these diverse models, we use the CNN
evaluator network to pick the best candidate from the pool of captions
generated by our top four models, \#1, \#2, \#4 and \#5.
The evaluator is trained using the gCNN+20Categ as the video feature.
It is shown as model \#6 in Table~\ref{tab:resultsVal}.
We can see that the CNN evaluator significantly outperforms, in all
the four metrics, every model it picks its candidates from.
Figure~\ref{fig:capSamps} shows an example in the third column where
the CNN evaluator picks a better caption than the one generated by our
best single model.


\subsection{Challenge Results}

%

Since the CNN evaluator model performed the best on the validation
set, we submitted that result in the MSR-VTT Challenge.
Our submission appears on the leaderboard as \emph{Aalto}.
The submissions were evaluated on the blind test set using the above
automatic metrics.
The results are shown in Table~\ref{tab:resultsTestMet}.
Our submission was ranked overall second considering the average ranking
across the metrics.

The submissions were also subject to human evaluation as the automatic
metrics are known to deviate from human judgements in case of both
image~\cite{CocoChallengeSlides} and
video~\cite{DBLP:journals/corr/RohrbachTRTPLCS16} data.
The human evaluation was based on three criteria: Coherence
(C1), Relevance (C2) and Helpfulness for the blind (C3).
Table~\ref{tab:resultsTestHum} presents the results of the human
evaluation.
As per human judgement, our submission was ranked the first among the
22 entries in the challenge.

Comparing the automatic metric and human evaluation based leaderboards
we see that the disagreement between the two is relatively minor, with
most teams in the top 10 changing their ranking by only one position.
This can most likely be attributed to having a large number of 20
reference captions per video for the automatic evaluation.


\section{Conclusions}
\label{sec:discussion}

We have presented our video captioning model which shared the first
position in the MSR Video to Language Challenge.
Our architecture consists of multiple caption generator models, based
on the encoder--decoder framework and trained on different video
features, and an evaluator model trained to pick the best candidate
from the ensemble.
We argue that this model is well-suited for the MSR-VTT dataset which
contains diverse types of videos.
Corroborating our claims, our model topped the human evaluation based
leaderboard and finished as second in the automatic metric based
leaderboard in the challenge.


\section{Acknowledgments}

This work was funded by the grant 251170 of the Academy of Finland and
calculations performed with resources of the Aalto University School
of Science ``Science-IT'' project.

\FloatBarrier

\bibliographystyle{abbrv}

\bibliography{../bibs/picsom,../bibs/others,../bibs/gicebib}

\end{document}